# Uncovering Key Trends in Industry 5.0 through Advanced AI Techniques

Panos Fitsilis

University of Thessaly, Department of Business Administration, pfitsilis@gmail.com

Paraskevi Tsoutsa

University of Thessaly, Department of Accounting and Finance, ptsoutsa@uth.gr

Vyron Damasiotis

University of Thessaly, Department of Accounting and Finance, ptsoutsa@uth.gr

Vasileios Kyriatzis

University of Thessaly, Department of Digital Systems, kyriatzis@uth.gr

This article analyzes around 200 online articles to identify trends within Industry 5.0 using artificial intelligence techniques. Specifically, it applies algorithms such as LDA, BERTopic, LSA, and K-means, in various configurations, to extract and compare the central themes present in the literature. The results reveal a convergence around a core set of themes while also highlighting that Industry 5.0 spans a wide range of topics. The study concludes that Industry 5.0, as an evolution of Industry 4.0, is a broad concept that lacks a clear definition, making it difficult to focus on and apply effectively. Therefore, for Industry 5.0 to be useful, it needs to be refined and more clearly defined. Furthermore, the findings demonstrate that well-known AI techniques can be effectively utilized for trend identification, particularly when the available literature is extensive and the subject matter lacks precise boundaries. This study showcases the potential of AI in extracting meaningful insights from large and diverse datasets, even in cases where the thematic structure of the domain is not clearly delineated.

## 1 INTRODUCTION

The term "Industry 5.0" represents the evolution of the well-established concept of Industry 4.0, reflecting a shift towards a more sustainable, human-centric, and resilient industrial paradigm. While Industry 4.0 focuses primarily on the efficiency, automation, and digitization of manufacturing processes, Industry 5.0 aims to broaden this vision. It not only emphasizes productivity but also incorporates the wellbeing of workers, societal prosperity, and the planet's ecological boundaries into the core of industrial strategies. This new vision for European industry aims to drive both digital and green transitions by placing technological innovation at the service of humanity and sustainability.

As highlighted by the European Commission, Industry 5.0 [1] promotes a future-oriented approach where industries actively contribute to solving societal challenges such as resource preservation, climate change, and social stability. By adopting circular production models and optimizing the use of natural resources, Industry 5.0 offers solutions that extend beyond traditional business objectives [2, 3]. It empowers workers by addressing evolving skill needs and training requirements, enhances industrial competitiveness, and attracts top talent. Moreover, revising existing value chains and improving energy efficiency allows industries to become more resilient to external shocks, such as the COVID-19 pandemic. In this context, Industry 5.0 not only benefits businesses but also supports the wider society, ensuring that technological progress serves human prosperity and environmental sustainability.

While the concept of Industry 5.0 presents a compelling future vision, it also faces significant challenges due to its broad and often ambiguous definition. Unlike Industry 4.0 [4, 5], which is closely associated with specific technologies like the Internet of Things (IoT), artificial intelligence (AI), and automation, the scope of Industry 5.0 is not as clearly delineated. This ambiguity has led to difficulties in determining which technologies and innovations truly belong to the realm of Industry 5.0, resulting in confusion among researchers and practitioners alike. Although there is a growing body of literature on the subject, it often lacks clear direction, making it challenging to develop a comprehensive understanding of the concept.

To address this issue, it is essential for research communities and industry stakeholders to work together to create taxonomies and frameworks that define Industry 5.0 more clearly. Such efforts would not only aid researchers in navigating the extensive and diverse literature but also provide businesses and organizations with roadmaps to implement Industry 5.0 strategies effectively. These taxonomies could serve as a foundation for establishing clear definitions and boundaries within Industry 5.0, guiding the future development and application of the technologies associated with this industrial revolution.

Artificial intelligence (AI) techniques offer a promising solution to this challenge by automating the analysis of large volumes of literature. With AI algorithms like Latent Dirichlet Allocation (LDA), BERTopic, Latent Semantic Analysis (LSA), or simpler algorithms such as K-means clustering etc. [6, 7, 8], researchers can automatically categorize academic and industrial publications into thematic clusters, identifying the central topics that define Industry 5.0. These algorithms allow for the extraction of patterns and trends from diverse and unstructured text data, helping to establish a more structured understanding of Industry 5.0. By analyzing the vast body of literature in this way, AI can assist in creating a roadmap for researchers and professionals, offering insights into the key areas of focus within the emerging landscape of Industry 5.0.

The objective of this article is precisely that—to leverage AI algorithms to uncover the key trends and themes that characterize Industry 5.0 as reflected in the literature. By applying various machine learning models, this study aims to provide an automated approach to understanding the core topics of Industry 5.0. The analysis will focus on classifying a collection of articles into meaningful categories, with the ultimate goal of guiding future research and industrial practices.

This article is structured as follows: Section 2 presents a review of the key literature on Industry 5.0 and provides a brief introduction to the AI algorithms used for topic modeling. Section 3 outlines the research methodology, detailing the steps involved in applying the selected AI techniques. Section 4 presents the results of the analysis, comparing the findings from different algorithms and highlighting the convergence of central themes. Finally, Section 5 discusses the conclusions, limitations, and future work planned in this area. By exploring the intersection of Industry 5.0 and advanced AI techniques, this article contributes to the growing understanding of this emerging industrial paradigm and offers insights into how AI can be used to advance the field.

## 2 BACKGROUND

### 2.1 Towards industry 5.0

Industry 5.0 represents a shift beyond the principles of Industry 4.0 by focusing on a more human-centric, resilient, and sustainable approach. It integrates contemporary technologies while at the same time is focusing on the role of people in driving value creation. Below we give a very brief overview on the key technologies, the social aspects, and economic considerations of Industry 5.0.

**Technologies of Industry 5.0**

Industry 5.0 builds on technologies that were established during the fourth industrial revolution but introduces significant advancements emphasizing sustainability. The key technologies are grouped into facilitating and emerging groups. Facilitating technologies stem from Industry 4.0, while emerging technologies are those aiming to bring sustainability and civic aspects in the domain of Industry 5.0. More specifically:

- Facilitating technologies, such as big data analytics, cloud computing, and enterprise systems, serve as fundamental building blocks of Industry 5.0. These technologies were pioneered in Industry 4.0 and continue to drive productivity and efficiency improvements. They are integral to the ongoing digital transformation of industrial systems, providing the foundation for more complex and interconnected technological advancements[9, 10].
- Emerging technologies in Industry 5.0 are designed to create eco-friendly and human-centric value-creation methods. Examples include Cognitive Cyber-Physical Systems (C-CCP), which enhance the human-machine interface by incorporating a level of machine consciousness. This system allows for proactive decision-making and improves man-machine interactions through pattern detection, failure correction, and decision-making[9, 10].

Other emerging technologies such as Cognitive Artificial Intelligence (CAI) integrate AI with human-like cognition. CAI supports decision-making, reduces information overload, and improves product sustainability by allowing machines to act and re-learn in human-like ways [9]. Human Interaction and Recognition Technologies (HIRT) further promote the human-centric ethos of Industry 5.0 by enhancing human-machine interaction in physical and cognitive tasks, making processes more efficient and user-friendly. Technologies like Extended Reality (XR), Industrial Smart Wearables (ISW), and Intelligent Robots are expected to play crucial roles in Industry 5.0. XR enables immersive industrial applications, while ISW provides workers with tools to enhance their performance, safety, and productivity. Adaptive robots can perform complex tasks alongside humans and are predicted to evolve significantly under Industry 5.0 [12].

**Social Aspects of Industry 5.0**

A distinguishing feature of Industry 5.0 is its focus on human-centered approaches, which strive for a balance between automation and creativity. The main aim is to empower people with tools and systems that enhance their productivity while ensuring sustainability and well-being [9]. More specifically, key aspects are:

- While Industry 4.0 prioritized automation, Industry 5.0 focuses on Human-Machine Collaboration, Cognitive AI and Cognitive Cyber-Physical Systems (C-CCP) aiming to complement human qualities like creativity, critical thinking with speed, stamina and precision of machines [2, 13].
- Ethical and sustainable business practices are valued Industry 5.0 and aligned with growth. Through technologies like Intelligent Energy Management Systems (IEMS) and Digital Twin (DSDT) technologies are promoted in Industry 5.0 since the lead to efficient use of resources, reduced energy consumption, and the development of a circular economy [14] .

**Economic Aspects of Industry 5.0**

The economic implications of Industry 5.0 extend beyond traditional measures of profitability and productivity, focusing on resilience, adaptability, and sustainability that are considered as "new" business requirements (or ESG factors).

- Industry 5.0 aims to create resilient industrial systems that can withstand stresses and disruptions. This can be achieved by utilizing strategies such as Smart Supply Chains and/or Smart Logistics, Industry 5.0 intends to ensure business continuity and fast recovery after unexpected disruptions. This implies that are available systems for managing real-time data and/or predictive analytics to manage risks, techniques for improving operational agility, etc. [15]
- A key advantage of Industry 5.0 is its focus on personalized products and services, facilitated by Smart Product Lifecycle Management (SPLM) systems and Dynamic Simulation and Digital Twin (DSDT) technologies. These strategies enable producers to offer customized products while lowering resource consumption and reducing waste [2].

In conclusion, Industry 5.0 marks a transformative stage in the evolution of industrial systems. It leverages advanced technologies while prioritizing ESG factors.

## 2.2 Algorithms for topic modeling

Topic modeling techniques are essential in text mining and natural language processing for identifying hidden themes in large datasets. One of the most popular methods is Latent Dirichlet Allocation (LDA), a probabilistic model that assigns words to topics based on their co-occurrence across documents [18, 19]]. LDA is widely used for its simplicity and effectiveness in capturing thematic structures. Another key approach is Latent Semantic Analysis (LSA) [20, 21], which applies singular value decomposition (SVD) to a term-document matrix, reducing the dimensionality and uncovering latent relationships between terms and documents. BERTopic is a more recent development that leverages transformer models and dynamic embeddings to generate high-quality, coherent topics [22]. BERTopic stands out for its ability to handle contextualized text representations and is particularly useful in cases where fine-tuning based on word context is needed.

Each topic modeling technique offers benefits and is appropriate for different types of datasets, depending on the structure, size, and complexity of the data. More specifically,

- LDA is highly interpretable and flexible, making it one of the most popular techniques. It based on Dirichlet probabilistic distribution of words across topics, making it easier to assign thematic meanings to clusters of words. LDA works well when the dataset consists of moderate-length texts (e.g., academic articles, news articles, or long-form reviews). It performs best when the assumption of topics being a mixture of words is valid, and it offers control over the number of topics. However, LDA might not perform well on short texts or contexts with semantics.
- LSA main benefit is the reduction of dimensionality through matrix factorization. This helps to uncover relationships between terms and documents that might not be immediately obvious. It is useful for identifying latent structures in data without requiring a probabilistic model. It is more effective when applied to datasets that are well-structured but contain noise or redundant information, as it can smooth out noise while highlighting key thematic connections. It is commonly used with medium to large datasets (e.g., corpora of academic papers or legal documents) and is suitable when data dimensionality reduction is needed. LSA, however, faces difficulties with ambiguity and lacks the ability to capture contextual nuances.
- BERTopic is particularly strong in handling contextualized text and short texts, leveraging advanced transformer-based embeddings (e.g., BERT) to capture the semantics of words in their specific context. This approach generates highly coherent topics and adapts dynamically to changing datasets. BERTopic is ideal for datasets with complex, nuanced language or contexts, such as social media posts, customer reviews, and conversational data, where understanding the meaning of words in context is crucial. Its ability to handle diverse, smaller datasets makes it particularly valuable for dynamic, real-time data like tweets or news headlines.

## 3 RESEARCH METHODOLOGY

The research methodology employed in this study aimed at uncovering key trends and themes in Industry 5.0, using advanced AI-based topic modeling algorithms on a comprehensive dataset of articles related to the subject. The methodology followed a systematic and structured approach, ensuring the reliability and validity of the results by leveraging different technological and analytical tools. This section outlines each step in detail, justifying the choices made throughout the research process.

**Step 1: Article Collection**

The first step involved the collection of a corpus of articles centered on Industry 5.0. The primary objective was to compile a comprehensive dataset from multiple reliable sources, ensuring diversity and inclusivity in perspectives and data sources. The selected articles covered the last five years, a timeframe chosen to capture the most recent developments in Industry 5.0 technologies and applications.

The initial search for relevant articles was conducted using Google Scholar, a well-established platform for academic research. The query focused on terms directly related to Industry 5.0. Given that Industry 5.0 is a relatively recent development, the search was limited to the last five years to ensure the inclusion of the latest advancements and trends. Following this, the 100 most relevant articles were selected based on their citation count, relevance to the theme, and availability. These criteria helped filter high-impact studies and articles contributing significant insights into Industry 5.0.

To further diversify the dataset and include non-academic perspectives, additional articles were collected from general Google and Microsoft Bing search results, focusing on white papers and industrial reports. Ensuring that the dataset contained articles from varied sources was critical. By collecting content from academic, industrial, and popular platforms, we were able to mitigate bias and provide a comprehensive understanding of Industry 5.0 from different stakeholders. This diversity helped in forming a more accurate portrayal of the state of research, development, and implementation of Industry 5.0 concepts across industries. By the end of this step, approximately 200 documents had been gathered, including a balanced mix of academic publications, industry reports, and popular literature. After a manual screening 181 documents were retained.

**Step 2: Data Preprocessing**

The second step involved preparing the dataset for analysis. Given that the data consisted of text-based PDF files, several preprocessing steps were taken to transform the raw documents into a format suitable for algorithmic analysis. The goal of this step was to reduce noise and remove irrelevant data while preserving the core content of each document.

a) All PDF files were converted to plain text format to facilitate easier manipulation and processing. This conversion was carried out using Python-based tools capable of handling large volumes of PDF documents. By converting the articles to text files, we were able to perform text preprocessing steps programmatically, ensuring consistency across the dataset.

b) Removal of Stopwords and Noise. The next preprocessing task was the removal of common stopwords—words like "and," "the," "of," etc., which do not contribute to the thematic content of the articles. The Python library NLTK (Natural Language Toolkit) was employed to perform this task efficiently. Additionally, a custom list of stopwords was created based on the most frequently occurring non-relevant words within the dataset. This custom stopword list was generated by analyzing the data using ChatGPT, which helped identify domain-specific terms and phrases that were irrelevant to the thematic analysis (e.g., publication-specific metadata like "figure," "table," and other frequently repeated sections like "references").

c) To further refine the dataset, we removed the references section from each document, as this section typically does not contribute to the main themes of the text. Given the academic nature of many of the documents, removing the reference lists was essential to prevent skewing the results toward citations rather than content.

**Step 3: Application of Topic Modeling Algorithms**

Once the dataset was preprocessed, it was ready for the application of various topic modeling algorithms. Topic modeling is a machine learning method used to identify patterns and thematic structures within large corpora of documents. In this research, several algorithms were applied, including Latent Dirichlet Allocation (LDA), BERTopic and Latent Semantic Analysis (LSA). Each algorithm was applied with different parameter configurations to optimize the performance and ensure the robustness of the results.

**Step 4: Guided Lists and Zero-Shot Learning**

To improve the model's performance and ensure the relevance of the topics discovered, we implemented two different strategies: guided lists and zero-shot learning.

a) For some of the experiments, a guided list was introduced to influence the model's output. The guided list was created using terms from the ESCO database, a well-established resource of skills and technology terminologies. This list included 1294 technology specific terms from the "DigitalSKillsCollection" dataset. related to Industry 5.0 technologies, like "big data," "smart manufacturing," and "human-machine collaboration." The guided list was fed into models like BERTopic, allowing them to focus on Industry 5.0-related terms, thereby improving the quality and specificity of the results.

b) In other cases, a zero-shot learning approach was used, where predefined popular terms related to Industry 5.0 generated from ChatGPT and analyzed. These terms were cross-referenced with the input data to identify frequently occurring concepts and emerging trends.

**Step 5: Evaluation of Results**

Finally, the results from the different models were evaluated having as a goal was to find the optimal number of topics and the best-performing model in terms of uncovering significant trends in Industry 5.0. We have not used quantitative methods since the same metrics was not available for all algorithms. The way each algorithm was used is presented in Table 1.

Table 1: Algorithm application parameters

| Algorithm | Usage logic |
| --- | --- |
| **K-Means** | Initially, TF-IDF was used to convert the preprocessed texts into numerical vectors. It calculates the importance of words in the corpus by multiplying the term frequency by the inverse document frequency. The vectorization parameters included a maximum document frequency (max_df=0.80), a minimum document frequency (min_df=3), and a custom stopword list. The vocabulary was restricted to a predefined seed list. Subsequently, K-Means was applied to the TF-IDF vectors (document-term matrix) to group documents into clusters (interpreted as topics). Hyperparameter tuning was performed using GridSearchCV to find the optimal number of clusters, initialization methods (random and k-means++), maximum iterations, number of runs with different centroid seeds (n_init), and tolerance for stopping criteria (tol). The best parameters were selected based on a cross-validation process. |
| **Latent Dirichlet Allocation (LDA)** | The CountVectorizer was used to transform the preprocessed texts into a document-term matrix based on word counts. It was configured with custom stopwords and an n-gram range of (1, 2), allowing for both single words and word pairs (bigrams) to be considered in the model. The resulting matrix was used as the input for the LDA model. The LDA model was configured to generate 5 topics, with the random state set for reproducibility. The model worked by assigning each document a distribution of topics and identifying the words most associated with each topic. After fitting the model, the resulting topics were transformed from the input texts. |
| **Latent Semantic Analysis (LSA)** | The TfidfVectorizer was used to transform the preprocessed texts into a document-term matrix based on the Term Frequency-Inverse Document Frequency (TF-IDF) metric. This vectorizer was configured with custom stopwords and an n-gram range of (1, 2), allowing for single words and bigrams to be included in the analysis. The resulting matrix served as input for the Latent Semantic Analysis (LSA) model. The LSA model, implemented via TruncatedSVD, was applied to the TF-IDF |

| Algorithm | Usage logic |
|---|---|
|  | matrix. The model was configured to reduce the matrix down to 5 components, which represented the latent topics. The algorithm used was 'arpack', and the random state was set for reproducibility. The TruncatedSVD model fitted on the document-term matrix was used to identify hidden topics in the dataset. |
| **BERTopic not guided** | The SentenceTransformer model (all-mpnet-base-v2) was used to generate dense vector embeddings for each of the preprocessed texts. This transformer-based model captures contextual meaning and semantic relationships between the words in the texts, allowing for a more accurate representation of the documents as numerical embeddings. These embeddings were used as input for the BERTopic model. A custom UMAP model was initialized to reduce the dimensionality of the text embeddings generated by the SentenceTransformer. The UMAP algorithm is configured with specific parameters (n_neighbors=15, min_dist=0.1, n_components=2, and cosine metric) to ensure that similar texts are placed closer together in the reduced space. The reduced dimensions facilitate better clustering of the texts into topics. BERTopic was used to cluster the document embeddings into topics. The calculate_probabilities parameter was set to True, enabling the model to compute the probability distribution over topics for each document, which provides more granular insights into the topic associations. After fitting the BERTopic model on the preprocessed texts, the fit_transform method was applied to assign each document to a topic. The model output included both the assigned topic for each document and the corresponding probability distribution across topics, allowing for a probabilistic understanding of the topic assignments. |
| **BERTopic guided** | Again the SentenceTransformer model (all-mpnet-base-v2) was employed. The zero_shot_topics list containing predefined topics related to Industry 5.0 and technical areas was passed to the BERTopic model. These predefined topics serve as seed topics, guiding the BERTopic model to map the documents to these topics without requiring manual labeling. The use of zero-shot topics ensures that the model considers these predefined categories during the topic clustering process. The BERTopic model was initialized with the SentenceTransformer embeddings, the custom UMAP dimensionality reduction, and the zero-shot topics list. BERTopic was configured to calculate probabilities for each topic, which allows a probabilistic distribution over topics for each document. This enables the identification of multiple topic associations for a given text. After fitting the BERTopic model on the preprocessed texts using the fit_transform method, the model returned the assigned topic for each document along with the probability distribution across topics. This probability-based output allowed for more nuanced topic categorization, indicating the strength of each document's association with different predefined topics. |

## 4 DISCUSSION AND FINDINGS

The main results from the application of the above algorithms is presented in Table 2.

Table 2: Algorithm application results

| Algorithm | Results |
|---|---|
| **K-Means** | The analysis using K-means clustering with hyperparameter tuning yielded seven distinct clusters, which is more than in subsequent cases due to the inclusion of hyperparameters to optimize performance. The clustering results, however, indicate weak separation between the topics, as evidenced by a low silhouette score (0.041) and a Davies-Bouldin score of 3.44. Despite these limitations, each cluster represents a set of cohesive topics:<br>• **Cluster 5**: The largest cluster, containing 48 documents, is centered around themes such as "engineering, performance, strategy, operation, and risk." This group highlights the intersection of technical and strategic management topics. |

| Algorithm | Results |
|---|---|
| | - **Cluster 1**: The second-largest cluster (47 documents) focuses on "device, product, architecture, healthcare, and engineering." This cluster combines topics related to both healthcare innovation and engineering practices.<br>- **Cluster 2**: This cluster, with a smaller number of documents, emphasizes "project, market, governance, and strategy," showcasing topics that revolve around business governance and market-oriented strategies.<br>- **Cluster 3**: Documents in this cluster focus on "theory, problem, cybernetics, and history," delving into more abstract and academic discussions surrounding technological and theoretical frameworks.<br>- **Cluster 4**: Topics here revolve around "product, customer, logistics, and safety," highlighting operational concerns related to product development, logistics, and consumer safety.<br>- **Cluster 0**: This smaller cluster focuses on "processing, logic, and quality," suggesting a more technical theme related to production processes and quality control.<br>- **Cluster 6**: The smallest cluster highlights "concept, architecture, and twin," likely referring to digital twin technologies and conceptual architectural topics.<br><br>The use of hyperparameters contributed to the identification of a higher number of clusters, though the significant overlap between clusters points to limitations in the model's ability to cleanly distinguish between topics, as indicated by the low evaluation scores. |
| **Latent Dirichlet Allocation (LDA)** | The analysis conducted using the LDA (Latent Dirichlet Allocation) algorithm generated five key topics, reflecting diverse aspects of industrial, technological, and engineering challenges.<br>1. **Topic 0** focuses on the industrial and business transformation brought about by AI and digital technologies. Key terms like "industrial," "business," "society," and "revolution" highlight the impact of Industry 4.0 on various sectors. The central theme revolves around resource management and societal change due to AI integration.<br>2. **Topic 1** emphasizes the intersection of industrial services with AI and IoT (Internet of Things). Terms such as "challenge," "architecture," and "engineering" suggest a strong focus on addressing technological challenges, particularly through architectural and engineering solutions within the IoT framework.<br>3. **Topic 2** revolves around image processing techniques and algorithms, potentially linked to technical papers on image detection. While this topic includes technical aspects of IoT, the combination of "image," "algorithm," and "IoT" makes it somewhat less coherent, pointing to a broad application of these technologies across different fields.<br>4. **Topic 3** centers on the industrial workforce and their needs, particularly in the context of AI and business design. Words like "worker," "challenge," "product design," and "chain" suggest a focus on integrating AI into business processes and product development, emphasizing how industrial workers are impacted by these changes.<br>5. **Topic 4** is clearly focused on materials science and thermal engineering, featuring terms like "material," "temperature," "oil," "thermal," and "engine." This topic is well-defined, with cohesive words pointing to research or applications in heat management, composites, and engine performance in industrial settings.<br><br>The overall model demonstrates a broad coverage of industrial and technological themes, though certain topics (e.g., Topic 2) show less coherence, which might indicate overlapping or less clearly defined subject areas. The relatively high perplexity score suggests that there could be room for refinement to improve topic distinctiveness. |
| **Latent Semantic** | The analysis presents the following key clusters, each representing distinct themes and observations:<br>1. **Topic -1 (Outliers: ICT, System, Data):** This cluster comprises 107 documents that don't neatly fit into other categories, indicating a diverse mix of topics. The top keywords such as "ICT," |

| Algorithm | Results |
|---|---|
| **Analysis (LSA)** | "system," "management," and "data" suggest a broad range of documents related to information and communication technologies, data management, and systems integration. These documents might represent a combination of unrelated or very specific topics within the broader digital or industrial context.<br>2. **Topic 0 (Industry 4.0, Automation, and Technologies):** This cluster contains 40 documents and revolves around Industry 4.0 and its related technologies, including automation, systems, and product development. The keywords like "industry," "development," "systems," and "technologies" highlight the core focus on the technological advancements driving Industry 4.0. This cluster likely covers the integration of new technologies into manufacturing and production processes, as well as discussions about how industries are transforming through automation.<br>3. **Topic 1 (Education, Learning, and Digital Skills):** With 19 documents, this cluster focuses on the intersection of education, learning, and the development of digital skills. The top keywords like "learning," "education," and "digital" suggest a focus on how education systems are evolving to equip individuals with the necessary digital competencies for modern industries. This is especially relevant as digitalization becomes integral to both industrial and everyday contexts.<br>4. **Topic 2 (Green Development and Sustainable Practices):** This cluster, consisting of 12 documents, emphasizes sustainability and green development. Keywords such as "sustainability," "environment," and "green" highlight the focus on environmental considerations and practices aimed at promoting eco-friendly industries and business operations. This reflects growing trends in adopting greener technologies and processes as industries align with sustainability goals.<br><br>These clusters reflect the broad scope of topics related to Industry 4.0, education for digital skills, sustainability, and a wide range of ICT applications, illustrating the diverse challenges and opportunities in modern industries and education. |
| **BERTopic not guided** | The analysis reveals several key clusters, each with distinct thematic focuses:<br>1. **Topic -1 (Outliers: Industrial, Service Design, etc):** This cluster consists of 92 documents that don't strongly align with other topics, indicating outliers. However, the top keywords suggest that these documents pertain to mixed or niche subjects such as industrial applications, service design, AI, IoT and web-related services or technologies in industrial contexts. This cluster might require further exploration to better define or separate sub-topics within.<br>2. **Topic 0 (Society, Industrial Skills, and Revolution):** With 41 documents, this cluster focuses on the evolving relationship between society and industrial skills, driven by technological changes like the industrial revolution. The need for new skills and business transformations is emphasized, with mentions of workers, products, and businesses. This reflects the societal shifts triggered by the rapid industrial changes and the demands for skill adaptation.<br>3. **Topic 1 (AI, Robotics, and Industrial Applications):** This cluster, containing 35 documents, addresses the integration of AI, robotics, and IoT in industrial settings. It discusses both the challenges and risks of human-robot interaction, as well as the technical management of these applications (e.g., Product Data Management - PDM). The focus here is on the practical use of these technologies in real-world industrial processes.<br>4. **Topic 2 (Supply Chain and Environmental Focus):** Comprising 13 documents, this cluster emphasizes sustainability in supply chains. The presence of terms like "cleaner," "supply chain," and "environmental" indicates a focus on making supply chains more sustainable and eco-friendly. The cluster also suggests an academic focus, with frequent references to "journals" and "articles."<br><br>This clustering reveals significant themes in industrial transformations, skill development, AI integration, and environmental sustainability, all key areas for both industry and academic exploration. |

| Algorithm | Results |
|---|---|
| **BERTopic guided** | The document's topic modeling analysis has identified several key clusters, each reflecting important areas of focus.<br>**Cluster 0** centers on the relationship between society and industrial skills, highlighting the impact of industrial transformations on labor and the need for skill development in response to technological shifts such as the industrial revolution. This cluster discusses how new business models and industrial concepts are driving societal changes, emphasizing the importance of workers' skills.<br>**Cluster 1** focuses on the application of **AI and robotics** in industrial environments. It covers the integration of **AI, IoT, and robotics**, addressing the challenges associated with these technologies, such as human-robot interaction and risk management. There is a practical angle to this cluster, with references to **Product Data Management (PDM)** and the real-world challenges of applying advanced technologies in industrial settings.<br>**Cluster 2** is about **supply chain management** with a strong emphasis on **environmental sustainability**. The keywords in this cluster suggest a focus on making supply chains cleaner and more eco-friendly, touching on strategic discussions about sustainability in logistics and operations.<br>Lastly, the analysis also points out that there is a large portion of documents that did not strongly fit into any of the defined topics, falling under an **outlier cluster (Cluster -1)**. This indicates the presence of documents covering mixed or niche subjects, which could potentially form additional clusters if explored further. |

The results of the different models—LDA, LSA, and BERTopic (guided and unguided)—highlight different strengths and limitations in uncovering trends in Industry 5.0. LDA identified five broad topics, such as industrial transformation and AI integration, but had challenges with topic coherence, especially in overlapping areas like image processing and IoT. LSA provided more distinct themes, emphasizing automation, digital skills, and sustainability, but a large number of outlier documents suggested difficulties in clearly defining some topic boundaries.

BERTopic (not guided) showed granularity, capturing key areas like societal impacts, AI applications, and supply chain sustainability. However, it also revealed a significant number of outlier documents, hinting at niche or mixed topics. The guided BERTopic, using predefined zero-shot topics, focused on AI integration, robotics, and supply chain sustainability, offering a clearer topic mapping, but it too had some documents that didn't strongly fit predefined topics. Overall, BERTopic offered more nuanced insights, while LSA excelled in distinguishing broader categories.

## 5 CONCLUSIONS

This study applied a variety of AI techniques, including LDA, LSA, and BERTopic, to analyze approximately 200 documents related to Industry 5.0. The dataset, composed of articles from both academic and industrial sources, also varied significantly in terms of content quality and focus, potentially impacting the results. The results indicate that while each algorithm uncovered relevant themes, BERTopic (guided and unguided) performed best in capturing more granular and contextual insights. LDA and LSA, although useful in identifying broader trends, were less capable of differentiating niche topics. Overall, the findings emphasize that Industry 5.0 spans a wide range of areas, including AI, sustainability, and workforce transformations. However, there is still a need for clearer definitions to focus the research.

Future research should focus on refining the definition of Industry 5.0 and as well the related topics. Further, the data should integrate more diverse datasets. Further, applying unsupervised AI methods to even broader collections of literature it is expected to offer additional insights.

## REFERENCES

[1] European Commission. (n.d.). *Industry 5.0*. Research and Innovation. Retrieved October 18, 2024, from https://research-and-innovation.ec.europa.eu/research-area/industrial-research-and-innovation/industry-50_en

[2] Ghobakhloo, M., Iranmanesh, M., Tseng, M. L., Grybauskas, A., Stefanini, A., & Amran, A. (2023). Behind the definition of Industry 5.0: A systematic review of technologies, principles, components, and values. *Journal of Industrial and Production Engineering*, *40*(6), 432-447.

[3] Maddikunta, P. K. R., Pham, Q. V., Prabadevi, B., Deepa, N., Dev, K., Gadekallu, T. R., ... & Liyanage, M. (2022). Industry 5.0: A survey on enabling technologies and potential applications. *Journal of industrial information integration*, *26*, 100257.

[4] Erboz, G. (2017). How to define industry 4.0: Main pillars of industry 4.0. *Managerial trends in the development of enterprises in globalization era*, *761*, 761-767.

[5] Fitsilis, P., Tsoutsa, P., & Gerogiannis, V. (2018). Industry 4.0: Required personnel competences. *Industry 4.0*, *3*(3), 130-133.

[6] Dalvi, A., Joshil, V., Warior[1], A., & Nair[1], D. An Analysis of Topic Modeling Approaches. *Innovations and Advances in Cognitive Systems: ICIACS 2024, Volume 2*, 150.

[7] Wijanto, M. C., Widiastuti, I., & Yong, H. S. (2024). Topic Modeling for Scientific Articles: Exploring Optimal Hyperparameter Tuning in BERT. *International Journal on Advanced Science, Engineering & Information Technology*, *14*(3).

[8] George, L., & Sumathy, P. (2023). An integrated clustering and BERT framework for improved topic modeling. *International Journal of Information Technology*, *15*(4), 2187-2195.

[9] Petrescu, M. G., Neacșa, A., Laudacescu, E., & Tănase, M. (2023). Energy in the Era of Industry 5.0—Opportunities and Risks. In *Industry 5.0: Creative and Innovative Organizations* (pp. 71-90). Cham: Springer International Publishing.

[10] AIOTI WG Standard. (2023). *IoT and edge computing EU-funded projects landscape* (Release 2.0). AIOTI. https://aioti.eu/wp-content/uploads/2023/12/AIOTI-Report-EU-funded-research-projects-landscape-IoT-and-Edge-computing-R2-Final.pdf

[11] Olaizola, I. G., Quartulli, M., Garcia, A., & Barandiaran, I. (2022). Artificial Intelligence from Industry 5.0 perspective: Is the Technology Ready to Meet the Challenge?. *Proceedings http://ceur-ws. org ISSN*, *1613*, 0073.

[12] Leng, J., Sha, W., Wang, B., Zheng, P., Zhuang, C., Liu, Q., ... & Wang, L. (2022). Industry 5.0: Prospect and retrospect. *Journal of Manufacturing Systems*, *65*, 279-295.

[13] Ordieres-Meré, J., Gutierrez, M., & Villalba-Díez, J. (2023). Toward the industry 5.0 paradigm: Increasing value creation through the robust integration of humans and machines. *Computers in Industry*, *150*, 103947.

[14] Mourtzis, D., Angelopoulos, J., & Panopoulos, N. (2022). Smart grids as product-service systems in the framework of energy 5.0-a state-of-the-art review. *Green Manufacturing Open*, *1*(1), 5.

[15] Ejjami, R., & Boussalham, K. (2024). Resilient Supply Chains in Industry 5.0: Leveraging AI for Predictive Maintenance and Risk Mitigation. *IJFMR-Int J Multidiscip Res [Internet]*, *6*(4).

[16] Asmussen, C. B., & Møller, C. (2019). Smart literature review: a practical topic modelling approach to exploratory literature review. *Journal of Big Data*, *6*(1), 1-18.

[17] Guler, N., Kirshner, S. N., & Vidgen, R. (2024). A literature review of artificial intelligence research in business and management using machine learning and ChatGPT. *Data and Information Management*, 100076.

[18] Jelodar, H., Wang, Y., Yuan, C., Feng, X., Jiang, X., Li, Y., & Zhao, L. (2019). Latent Dirichlet allocation (LDA) and topic modeling: models, applications, a survey. *Multimedia tools and applications*, *78*, 15169-15211.

[19] Blei, D. M., Ng, A. Y., & Jordan, M. I. (2003). Latent dirichlet allocation. *Journal of machine Learning*

*research*, *3*(Jan), 993-1022.

[20] Dumais, S. T. (2004). Latent semantic analysis. *Annual Review of Information Science and Technology (ARIST)*, *38*, 189-230.

[21] Evangelopoulos, N. E. (2013). Latent semantic analysis. *Wiley Interdisciplinary Reviews: Cognitive Science*, *4*(6), 683-692.

[22] Atagün, E., Hartoka, B., & Albayrak, A. (2021, September). Topic modeling using LDA and BERT techniques: Teknofest example. In *2021 6th International Conference on Computer Science and Engineering (UBMK)* (pp. 660-664). IEEE.